# Scalable Medication Extraction and Discontinuation Identification from Electronic Health Records Using Large Language Models

Chong Shao[1,2], Douglas Snyder[2], Chiran Li[2], Bowen Gu[1], Kerry Ngan[1], Chun-Ting Yang[1], Jiageng Wu[1], Richard Wyss[1], Kueiyu Joshua Lin[1,#], Jie Yang[1,3,4,#]

[1] Division of Pharmacoepidemiology and Pharmacoeconomics, Department of Medicine, Brigham and Women's Hospital, Harvard Medical School, Boston, MA, USA

[2] Harvard T.H. Chan School of Public Health, Harvard University, Boston, MA, USA

[3] Broad Institute of MIT and Harvard, Cambridge, MA, USA

[4] Harvard Data Science Initiative, Harvard University, Cambridge, MA, USA

[#] **Co-senior authorship.**

***Corresponding author**: Jie Yang, PhD (jyang66@bwh.harvard.edu) and Kueiyu Joshua Lin, MD, ScD (jklin@bwh.harvard.edu), Division of Pharmacoepidemiology and Pharmacoeconomics, Brigham and Women's Hospital & Harvard Medical School, 75 Francis St, Boston MA 02115, USA,

**Main Text Word Count:** 2956

**Figures Count:** 5

**Tables Count:** 5

## Abstract

**Objective:** Identifying medication discontinuations in electronic health records (EHRs) is vital for patient safety but is often hindered by information being buried in unstructured notes. This study aims to evaluate the capabilities of advanced open-sourced and proprietary large language models (LLMs) in extracting medications and classifying their medication status from EHR notes, focusing on their scalability for medication information extraction without human annotation.

**Study Design and Setting:** We collected three EHR datasets from diverse sources to build the evaluation benchmark: one publicly available dataset (Re-CASI), one we annotated based on public MIMIC notes (MIV-Med), and one internally annotated on clinical notes from Mass General Brigham (MGB-Med). We evaluated 12 advanced LLMs, including general-domain open-sourced models (e.g., Llama-3.1-70B-Instruct, Qwen2.5-72B-Instruct), medical-specific models (e.g., MeLLaMA-70B-chat), and a proprietary model (GPT-4o). We explored multiple LLM prompting strategies, including zero-shot, 5-shot, and Chain-of-Thought (CoT) approaches. Performance on medication extraction, medication status classification, and their joint task (extraction then classification) was systematically compared across all experiments.

**Results:** LLMs showed promising performance on medication extraction, while discontinuation classification and joint tasks were more challenging. GPT-4o consistently achieved the highest average F1 scores in all tasks under zero-shot setting - 94.0% for medication extraction, 78.1% for discontinuation classification, and 72.7% for the joint task. Open-sourced models followed closely, with Llama-3.1-70B-Instruct achieving the highest performance in medication status classification on the MIV-Med dataset (68.7%) and in the joint task on both the Re-CASI (76.2%) and MIV-Med (60.2%) datasets. Medical-specific LLMs demonstrated lower performance compared to advanced general-domain LLMs. Few-shot learning generally improved performance, while CoT reasoning showed inconsistent gains. Notably, open-sourced models occasionally surpassed GPT-4o performance, underscoring their potential in privacy-sensitive clinical research.

**Conclusion:** LLMs demonstrate strong potential for medication extraction and discontinuation identification on EHR notes, with open-sourced models offering scalable alternatives to proprietary systems and few-shot learning further improving LLMs' capability.

**Plain Language Summary:** Stopping a medicine can affect safety and treatment decisions, yet this detail is often buried in long electronic health record notes. We evaluated whether large language models, which read and summarize text, can automatically find medication names and decide whether each medicine is still being taken, has been stopped, or neither. We tested 12 models, including open-source options suitable for secure hospital use, on three collections of clinical notes and compared three simple instruction styles: giving no examples, showing a few examples, and asking for step-by-step reasoning. All models produced usable results. The strongest systems scored about 94 for finding medication names and about 78 for deciding continued or stopped status, on a standard 0 to 100 measure that balances completeness and correctness. Showing a few examples usually helped more than step-by-step prompts, and several open-source models performed close to a leading proprietary system. These tools could help hospitals and researchers monitor medications at scale to support drug-safety studies, adherence tracking, and clinical decision support, with local validation and safeguards before clinical use.

**Keywords:** large language model; medication discontinuation; electronic health records, information extraction

Running Title: Medication Status Extraction from EHRs Using LLMs

## 1.0 Introduction

Accurately identifying medication discontinuation is essential for understanding adverse effects, therapeutic failures, and non-adherence[1], and for optimizing post-discharge care and informing clinical trial outcomes[2]. More broadly, medication-related information is critical for patient care, adherence monitoring, and drug safety surveillance[3–5]. While insurance claims data reflect prescription fills, they often miss non-reimbursed[6] medications or provider- or patient-driven discontinuations. Electronic health record (EHR) notes document these decisions and provide context on medication status. However, much is embedded in unstructured EHR notes[7], hindering large-scale analysis. Manual status annotation is resource-intensive, time-consuming, and not scalable. Thus, efficient natural language processing (NLP) is needed to extract medications and classify discontinuation status at scale, enabling large-cohort analyses for pharmacoepidemiology, drug safety/effectiveness, and healthcare quality improvement[8,9].

This study systematically benchmarks state-of-the-art LLMs for medication extraction and status classification. Specifically, we evaluate general-domain open-source model families (Llama 3.1, Qwen 2.5, Mistral), medical-specific models (Me-LLaMA and Meditron), and the proprietary GPT-4o for extracting medication mentions and classifying status (active, discontinued, or neither) from annotated clinical datasets[10–16]. Beyond an existing annotated dataset, we annotated two additional datasets, one from the MIMIC-IV[17,18] and one from an internal EHR source, to broaden evaluation. A key focus is detecting medication discontinuation status, a task with direct implications for clinical decision support and pharmacoepidemiologic research. Assessing performance across configurations identifies strengths and limitations and informs more effective uses of LLMs in clinical informatics.

## Statement of significance

| Problem or Issue | Accurately identifying medication discontinuation is essential for clinical decision-making, yet it is challenging due to the unstructured nature of electronic health records (EHRs). Manual annotation is resource-intensive and not scalable. |
|---|---|
| **What is Already Known** | Prior research has explored medication extraction from clinical text using traditional natural language processing (NLP) and machine learning methods, with limited evaluation of state-of-the-art large language models (LLMs). |
| **What this Paper Adds** | This study systematically benchmarks general-domain and medical-specific LLMs, including proprietary and open-source models, on medication extraction and medication status classification across diverse datasets. It also introduces novel annotated resources and focuses on evaluating performance in detecting medication discontinuation—an understudied but clinically significant task. |
| **Who would benefit from the new knowledge in this paper** | Clinical informatics researchers, health systems implementing EHR-based clinical decision support, and developers of NLP tools for drug safety, adherence monitoring, and pharmacoepidemiology. |

## 2.0 Related Work

Medication extraction and status classification remain challenging given diverse linguistic structures, implicit negations, and contextual nuances in clinical notes[19–21]. NLP for EHRs evolved from rule-based systems to statistical machine learning (e.g., logistic regression[22], XGBoost[23]) and supervised neural network models (e.g., convolutional neural networks[24]), and more recently, to large language models (LLMs). Rule-based approaches with expert patterns and dictionaries were widely used for medication extraction[25,26], but often lacked generalization across institutions and specialties[27]. Statistical machine learning and supervised neural network models improved scalability and performance for medication entity recognition and classification[28,29], yet require large, manually annotated training data, limiting cross-domain adaptation[30]. Mahajan et al constructed the Contextualized Medication Event Dataset (CMED)[31] that focuses on the mentioned-level medication-change status (i.e., Disposition, No Disposition, Undetermined). A feature-based SVM baseline achieved a macro-F1 of ~0.63 under cross-validation on the training split, and in the subsequent 2022 n2c2 shared task[32], the organizers reported a Conditional Random Field (CRF) baseline with an F1 of 0.746 on medication extraction.

## 3.0 Methods

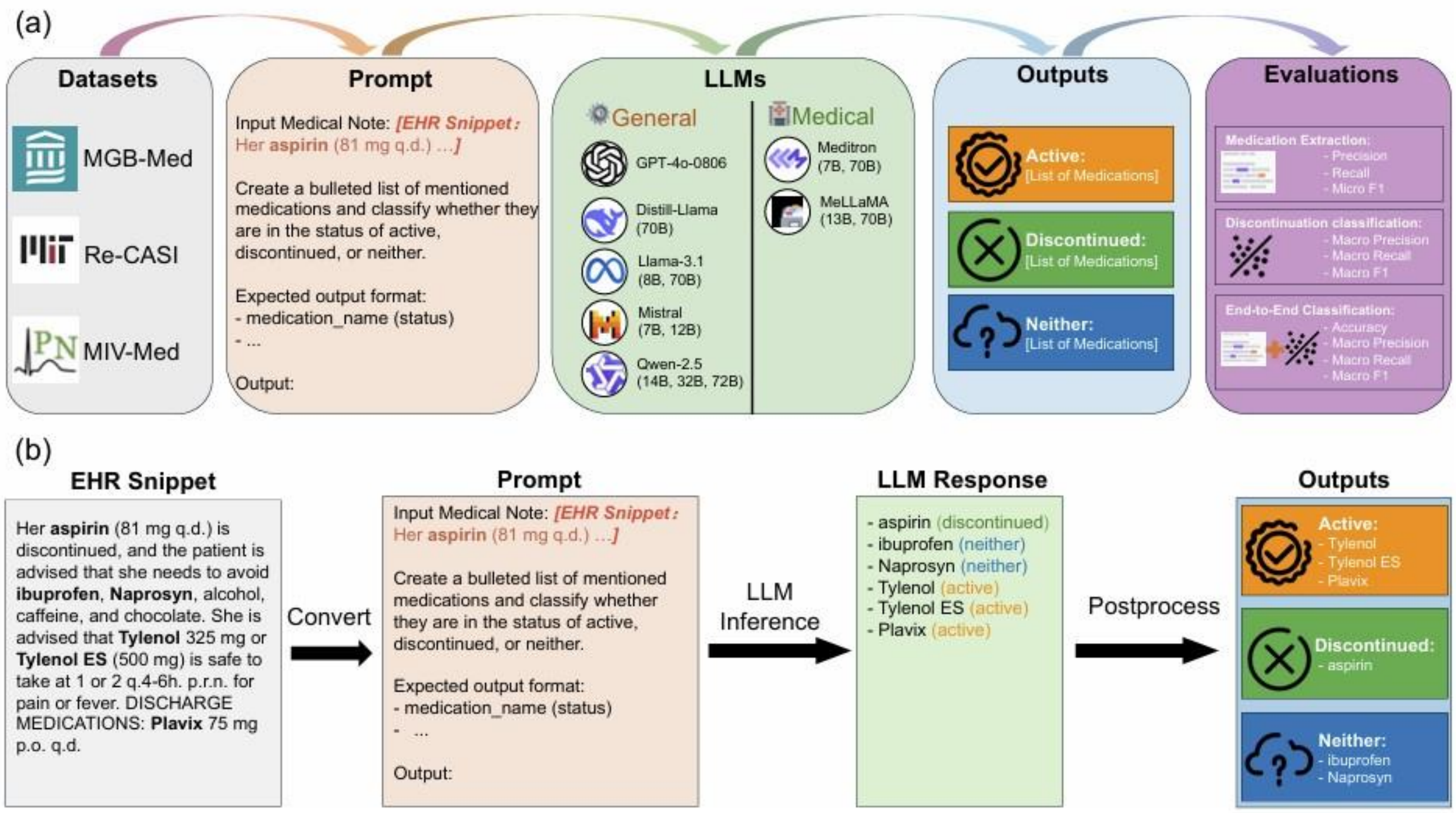

**Figure 1: Overview of Methods and Example Workflow** (a) Diagram of the methodology, illustrating the use of clinical text snippets from annotated datasets, processed through prompts and LLMs, with outputs evaluated using standardized metrics. (b) Example workflow showcasing how

a specific snippet is processed, from input through prompting and LLM inference to classification of discontinuation status.

### 3.1 Task overview

This study evaluates LLMs across three complementary tasks—medication identification, isolated status classification, and end-to-end classification (Figure 1).

**Medication Extraction:** Models detect and extract medication mentions from clinical narratives, evaluating correct identification of names.

**Medication Status Classification (Isolated Evaluation):** Given gold medication mentions, models classify each as "active" (patient is currently taking the medication), "discontinued" (the patient was previously taking the medication but has since stopped, as inferred from explicit statements or clear context in the clinical note), or "neither," (mentioned without evidence of current use or a stop event—e.g., historical/comparative mentions, allergy/"avoid" notes, or discussed-but-not-prescribed items) based on the note context (Appendix A.1). This isolates classification to measure contextual interpretation without extraction errors.

**End-to-End Classification (Joint task):** Models first extract medication mentions and then classify their discontinuation status. This joint setting reflects deployment, where errors propagate across stages (e.g., missed/extra entities affect status counts), thereby providing an overall measure of system utility on unstructured notes.

### 3.2 Datasets

**Table 1.** Dataset Characteristics

| Datasets | EHR Samples | Average Number of Words/Snippet | Medications Mentioned | Medication Continuation Status | | | Imbalanced Ratio[1] |
|---|---|---|---|---|---|---|---|
| | | | | Active | Discontinued | Neither | |

[1] Imbalance ratio (IR) = largest class count ÷ smallest class count within a dataset. With three classes, we report a single IR; per-class counts and percentages are shown in the table.

| MGB-Med | 241 | 90.93 | 443 | 327 | 116 | - | 2.8:1 |
|---|---|---|---|---|---|---|---|
| MIV-Med | 297 | 67.55 | 832 | 620 | 169 | 43 | 14.4:1 |
| Re-CASI | 100 | 62.61 | 336 | 190 | 112 | 34 | 5.6:1 |

We used three datasets to evaluate LLM performance: the internally curated MGB-Med dataset[33], the publicly available CASI dataset (Re-CASI)[16,34], and the re-annotated MIMIC-IV Medication Snippet dataset (MIV-Med)[17,18]. Details are in Table 1.

**MGB-Med Dataset**: An internal dataset from Brigham and Women's Hospital and Massachusetts General Hospital containing 241 human-reviewed EHR notes with machine-annotated medication mentions and status classification[33]. Snippets were extracted using tokenization, regular expressions, and rule-based matching, incorporating contextual sentences to enhance classification accuracy. Compared to other datasets, MGB-Med features longer and more detailed discharge summaries with medication directives embedded within broader clinical narratives. Unlike the other datasets, it omits a "Neither" category, simplifying classification but limiting coverage of ambiguous medication mentions.

**Re-Annotated CASI Dataset (Re-CASI)**: A publicly available, de-identified dataset containing 105 labeled examples derived from the Clinical Acronym Sense Inventory (CASI) dataset[16]. Of these, 5 examples were reserved exclusively for use in few-shot prompting, leaving 100 examples for evaluation, as in Table 1. It includes manually re-annotated medication mentions and status classifications (active, discontinued, or neither), providing a benchmark for structured prediction tasks.

**MIMIC-IV Medication Snippet Dataset (MIV-Med)**: Created from MIMIC-IV discharge summaries[17,18]. Medication-related snippets were identified using RxNorm-based[35] fuzzy matching, with two preceding and two following sentences added for context. Two annotators with medical terminology expertise (one MD) independently annotated 297 snippets as active, discontinued, or neither. The annotation process followed guidelines adapted from Re-CASI dataset (Appendix A.1) to ensure consistency. Inter-annotator agreement was strong, with a Cohen's κ=0.92 (Appendix A.2).

### 3.3 Models

We evaluated 12 advanced LLMs spanning sizes and domains:

- **Open-source general domain LLMs:** Llama-3.1-Instruct (8B, 70B)[10], Qwen-2.5-Instruct (14B, 32B, 72B)[12], Mistral models (Mistral-7B-Instruct-v0.3 (7B), Mistral-Nemo-Instruct-2407 (12B))[15].
- **Open-source medical-specific LLMs:** Me-LLaMA-chat (13B, 70B)[13], Meditron (7B, 70B)[14].
- **Proprietary LLM:** GPT-4o-0806[11], accessed via the MGB Azure platform to ensure EHR privacy.

### 3.4 LLM Inference Strategies

Three inference strategies (zero-shot, 5-shot, and CoT) were employed to evaluate LLM performance in extracting medication mentions and classifying their statuses, varying the number of examples and reasoning complexity. Prompts were adapted from a prior study[16] and customized for each dataset to match its label set. Specifically, the zero-shot prompt included no examples, while the 5-shot prompt contained five randomly selected examples. As the task context and annotation schemas matched, the same five examples from the Re-CASI dataset were reused for MIV-Med, and five reformatted examples were used for MGB-Med. The publicly accessible examples facilitate reproducibility, consistency, and clear illustration within the manuscript. The CoT prompt included the instruction "Let's think step by step" to promote stepwise reasoning before reaching the final answer. This widely used CoT approach has been validated across domains and in recent medical NLP studies[16,36,37], and this minimal prompt design isolates the effect of reasoning. Example prompts are in Appendix A.3.

### 3.5 Experiment Settings

Each experimental configuration was evaluated using five independent simulations. To introduce controlled randomness with reproducibility, we applied a fixed random seed (42) along with controlled stochastic generation parameters, including a temperature of 0.1 and a top-p of 0.9, while all other parameters were kept at their default values. Model outputs that do not follow the format instructions are treated as invalid responses.

This study was approved by the Mass General Brigham Institutional Review Board (IRB). Evaluations and analyses on the MGB-Med dataset were conducted on servers within the MGB network. Re-CASI and MIV-Med do not include protected health information (PHI); therefore, no PHI left the secure system.

### 3.6 Statistical Analysis

All results are presented as mean ± one standard deviation, computed across metric scores from the 5 independent runs. Each simulation utilized the complete dataset and applied the same generation parameters, ensuring variability reflected from the stochastic sampling.

### 3.7 Evaluations

Model performance was primarily assessed against human-annotated ground truth with macro- and micro-averaged F1 scores, as they balance precision and recall. Because label spaces differ, macro-F1 is not directly comparable across datasets. We interpret results within each dataset and do not aggregate F1 across datasets.

- **Medication Extraction:** Micro-F1 evaluates extraction accuracy. Precision and recall were reported for further analysis.

- **Medication Status Classification:** Macro-F1 was the primary metric, with equal weighting across status categories (“active,” “discontinued,” or “neither”). Precision and recall were also recorded to highlight class-specific performance.
- **End-to-End Classification (Joint Evaluation):** The main evaluation metric was macro-averaged F1-score, with accuracy reported to reflect overall correctness. Precision and recall were also included for a more detailed breakdown.

# 4.0 Results

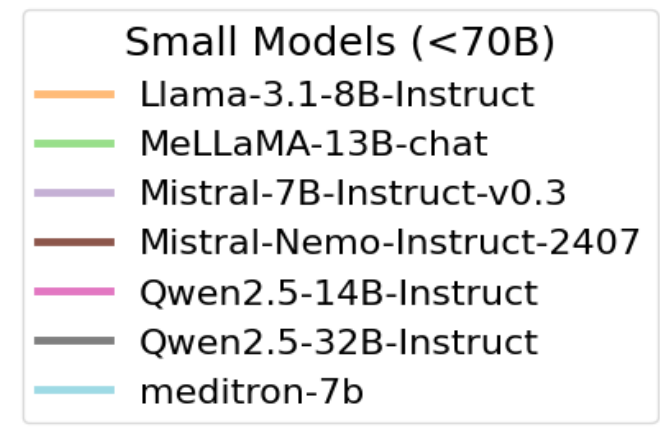

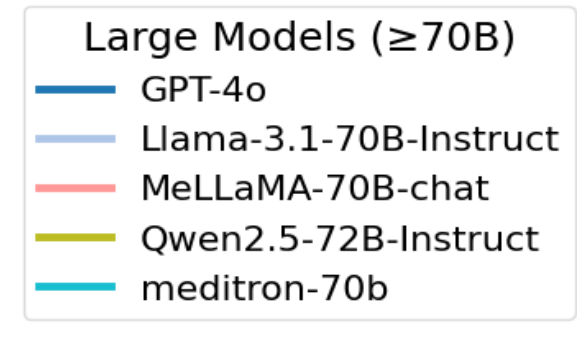

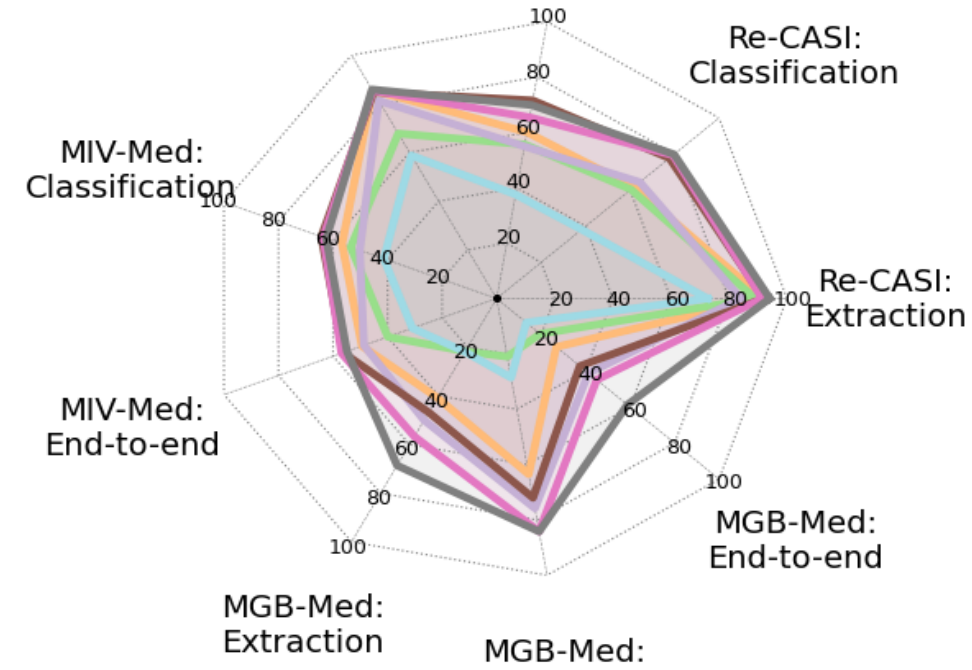

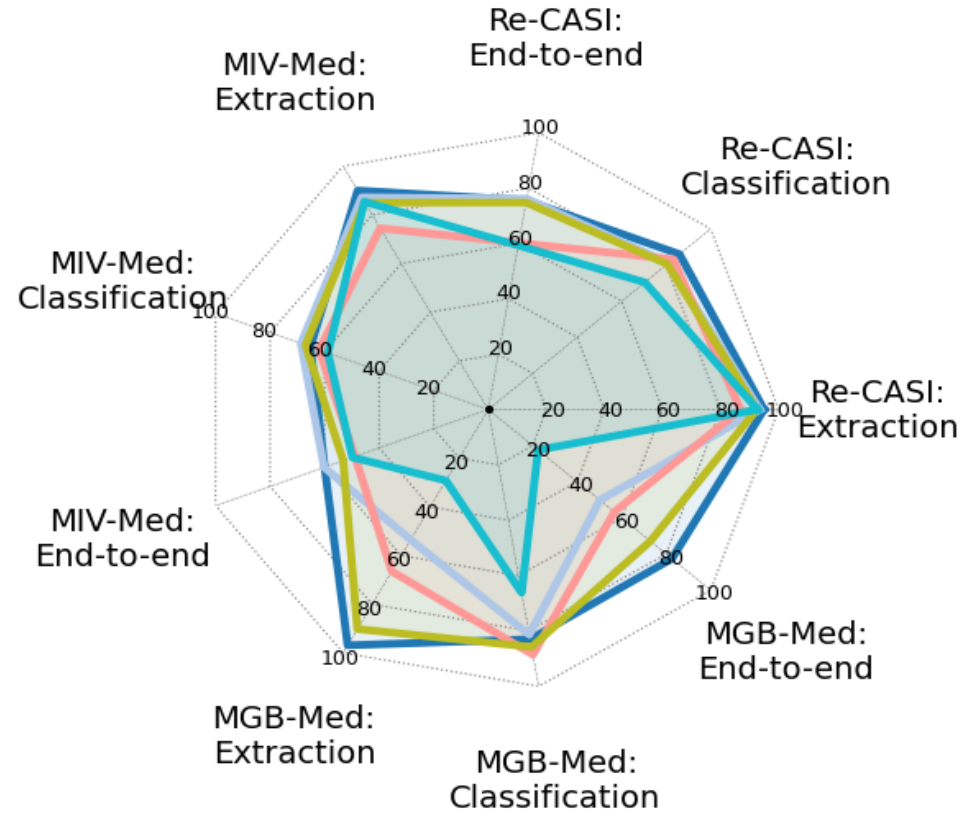

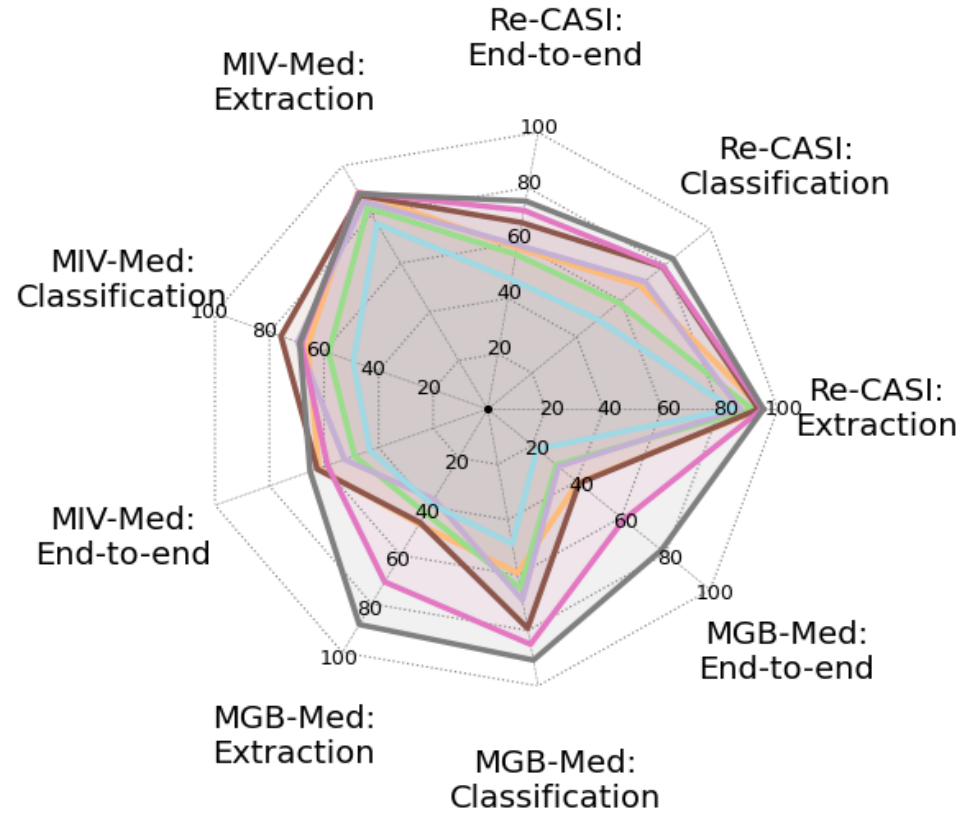

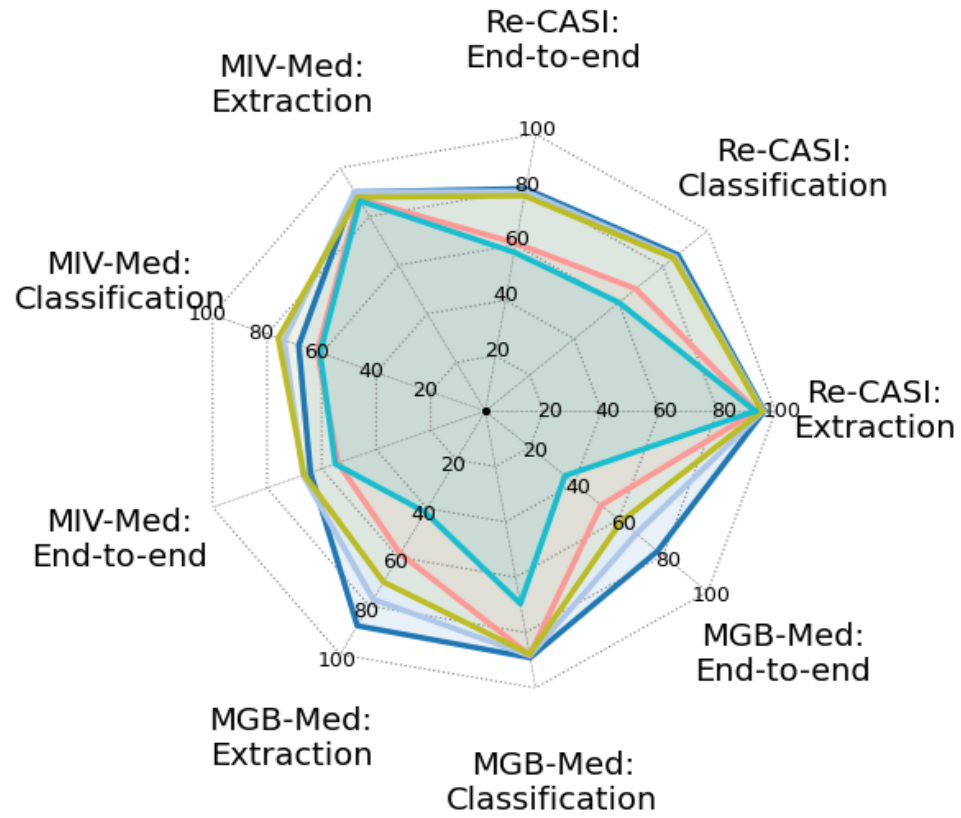

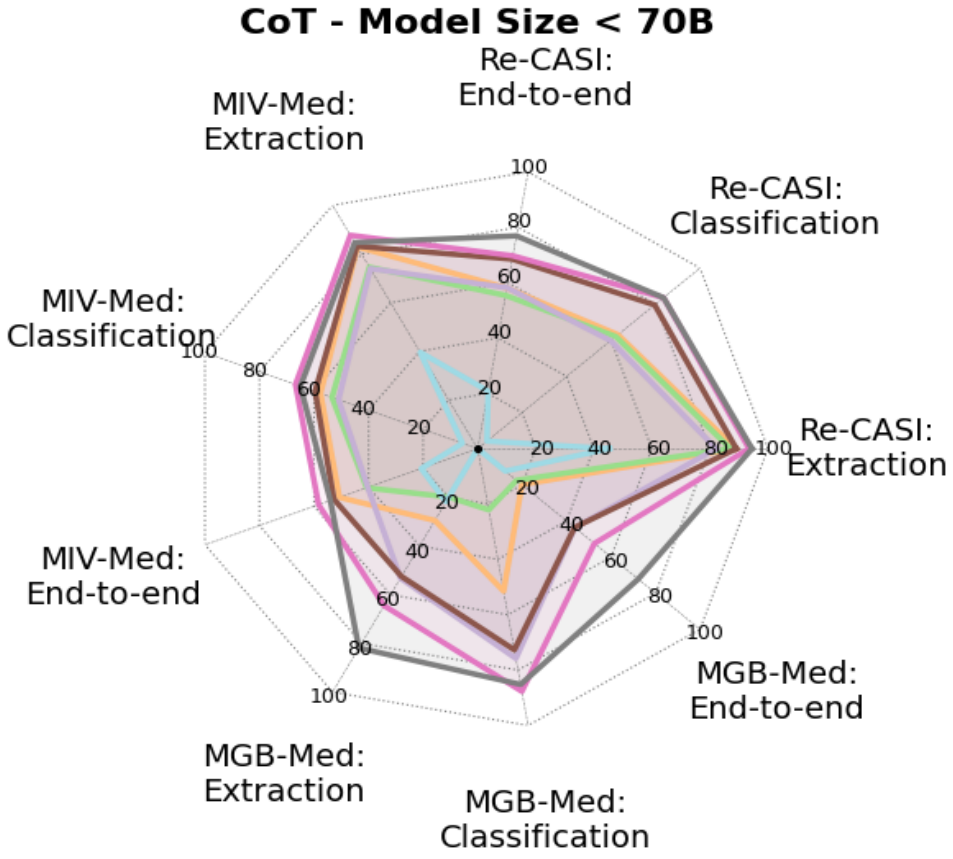

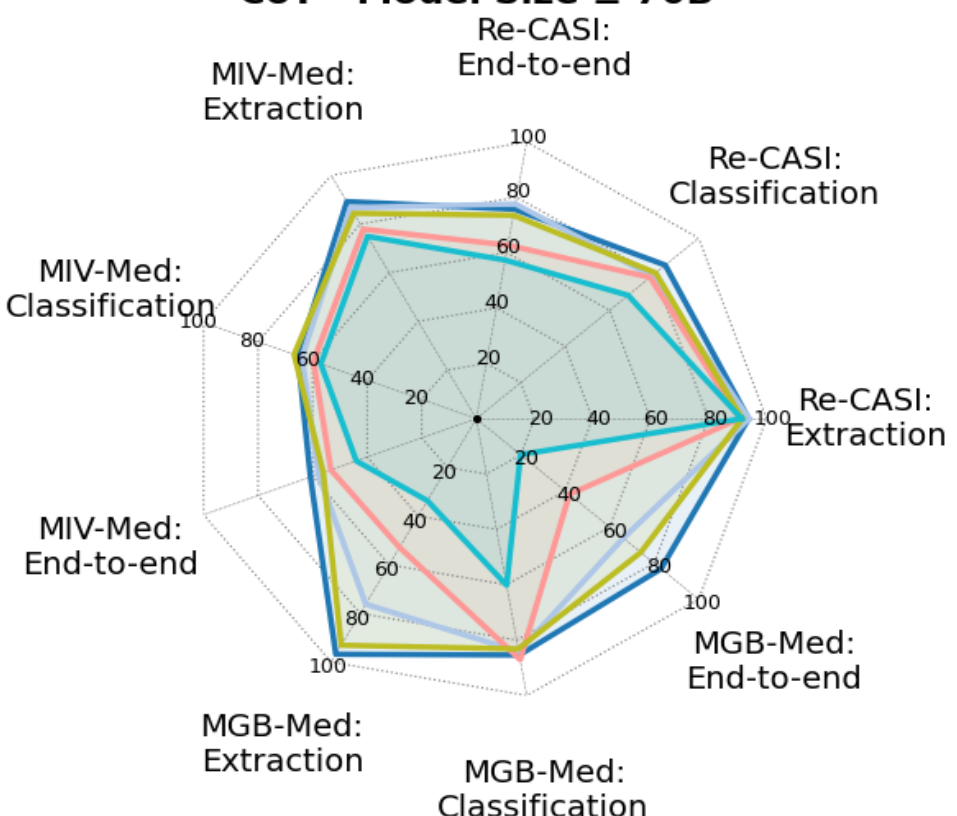

**Figure 2:** Model Performance Across Datasets and Inferences. Models with < 70B parameters are shown on the left, while those with ≥ 70B parameters are displayed on the right. Metrics used are F1-micro for extraction and F1-macro for classification and end-to-end experiments.

### 4.1 General performance

Figure 2 visualizes the performance of LLMs across datasets and tasks. Any "average across datasets" reported below is an unweighted descriptive mean for orientation only; conclusions rely on per-dataset results because label spaces differ. GPT-4o performed strongly across all tasks and prompting strategies, frequently ranking near the top within each dataset. Under zero-shot inference, GPT-4o achieved an average F1 of 94.0% for extraction and 78.1% for status classification, and 72.7% in the end-to-end setting. Qwen2.5-72B-Instruct and Llama-3.1-70B-Instruct also performed strongly among open-source models, slightly below GPT-4o overall but outperforming it in some settings. For example, Llama-3.1-70B-Instruct achieved the best discontinuation classification performance on the MIV-Med dataset in the zero-shot setting, while Qwen2.5-72B-Instruct obtained the highest average F1 score across the three datasets in the 5-shot setting. Tables 2–4 show GPT-4o was the most robust model across individual and combined tasks, and prompt strategies.

LLMs handled the extraction task relatively well. GPT-4o achieved an average F1 of 94.0%, while Qwen2.5-72B-Instruct reached 89.2% under zero-shot inference (Table 2). In contrast, medication status classification was more challenging: GPT-4o achieved 78.1%, followed closely by MeLLaMA-70B-Chat (77.9%) and Qwen2.5-72B-Instruct (77.8%), all under zero-shot settings. As expected with error propagation, end-to-end performance was lower; GPT-4o reached 72.7% (first), followed by Qwen2.5-72B-Instruct (second, 67.0%) and Llama-3.1-70B-Instruct (third, 62.2%).

**Table 2.** Model Performance under Zero-shot Prompting

| Model | Medication Extraction (%) (F1-micro) | | | | Medication Status Classification (%) (F1-macro) | | | | End-to-End (%) (F1-macro) | | | |
|---|---|---|---|---|---|---|---|---|---|---|---|---|
| | MGB-Med | Re-CASI | MIV-Med | Avg | MGB-Med | Re-CASI | MIV-Med | Avg | MGB-Med | Re-CASI | MIV-Med | Avg |
| GPT-4o | **96.9 ± 0.2** | **95.0 ± 0.2** | **90.0 ± 0.1** | **94.0** | 83.0 ± 0.4 | **86.0 ± 0.5** | 65.2 ± 0.7 | **78.1** | **82.6 ± 0.8** | 75.6 ± 0.7 | 60.0 ± 0.5 | **72.7** |
| Llama-3.1-8B-Instruct | 40.8 ± 0.6 | 92.0 ± 0.3 | 85.6 ± 0.3 | 72.8 | 63.5 ± 1.3 | 61.7 ± 0.7 | 56.7 ± 1.5 | 60.6 | 26.8 ± 0.5 | 60.1 ± 1.6 | 49.4 ± 1.5 | 45.4 |
| Llama-3.1-70B-Instruct | 53.5 ± 0.4 | 93.5 ± 0.2 | 87.2 ± 0.1 | 78.1 | 81.4 ± 0.7 | 81.1 ± 1.7 | **68.7 ± 0.5** | 77.1 | 50.2 ± 0.4 | **76.2 ± 0.6** | **60.2 ± 0.3** | 62.2 |
| Meditron-7b | 21.9 ± 0.0 | 72.7 ± 1.3 | 58.8 ± 1.4 | 51.1 | 28.6 ± 2.1 | 38.2 ± 0.7 | 41.4 ± 0.6 | 36.1 | 13.3 ± 0.5 | 37.6 ± 1.3 | 30.9 ± 1.5 | 27.3 |
| Meditron-70b | 29.2 ± 0.5 | 92.8 ± 0.4 | 85.4 ± 0.7 | 69.1 | 66.2 ± 2.1 | 70.4 ± 0.6 | 58.8 ± 0.2 | 65.1 | 22.4 ± 0.7 | 59.3 ± 1.3 | 49.8 ± 1.9 | 43.8 |
| MeLLaMA-13B-chat | 21.7 ± 0.0 | 89.1 ± 0.7 | 67.8 ± 0.6 | 59.5 | 21.0 ± 0.2 | 60.4 ± 0.7 | 53.5 ± 0.5 | 45.0 | 19.6 ± 0.2 | 55.7 ± 0.7 | 39.9 ± 0.3 | 38.4 |
| MeLLaMA-70B-chat | 66.5 ± 1.1 | 86.9 ± 0.6 | 74.5 ± 0.8 | 76.0 | **88.6 ± 0.2** | 83.2 ± 0.4 | 62.0 ± 0.6 | 77.9 | 56.7 ± 0.8 | 60.2 ± 1.2 | 48.9 ± 0.8 | 55.3 |
| Mistral-7B-Instruct-v0.3 | 50.0 ± 0.6 | 84.1 ± 0.3 | 81.3 ± 0.4 | 71.8 | 75.8 ± 0.4 | 64.4 ± 0.4 | 50.2 ± 0.6 | 63.5 | 41.3 ± 0.3 | 55.0 ± 0.2 | 48.7 ± 0.2 | 48.3 |
| Mistral-Nemo-Instruct-2407 | 46.8 ± 0.9 | 92.0 ± 0.5 | 84.8 ± 0.4 | 74.5 | 72.0 ± 1.1 | 77.7 ± 0.5 | 64.3 ± 0.3 | 71.3 | 37.4 ± 0.8 | 71.6 ± 1.0 | 56.3 ± 1.1 | 55.1 |
| Qwen2.5-14B-Instruct | 56.7 ± 1.0 | 91.5 ± 0.2 | 85.4 ± 0.8 | 77.9 | 84.2 ± 1.1 | 79.3 ± 1.0 | 63.6 ± 0.8 | 75.7 | 45.0 ± 0.8 | 65.5 ± 1.5 | 56.9 ± 0.5 | 55.8 |
| Qwen2.5-32B-Instruct | 68.6 ± 1.2 | 94.1 ± 0.1 | 85.9 ± 0.2 | 82.9 | 84.2 ± 1.0 | 79.8 ± 0.7 | 62.5 ± 0.2 | 75.5 | 58.8 ± 1.2 | 69.9 ± 0.8 | 54.9 ± 0.8 | 61.2 |
| Qwen2.5-72B-Instruct | 90.2 ± 1.3 | 92.5 ± 0.1 | 85.0 ± 0.1 | 89.2 | 85.8 ± 0.0 | 80.5 ± 0.7 | 67.0 ± 0.5 | 77.8 | 72.9 ± 1.7 | 74.8 ± 0.2 | 53.2 ± 0.3 | 67.0 |

*The models are listed in ascending alphabetical order and then by increasing parameter size. Bold text highlights the highest value in each column. The same formatting is applied in Tables 3 and 4.*

**Table 3**. Model Performance under 5-Shot Prompting

| Model | Medication Extraction (%) (F1-micro) | | | | Medication Status Classification (%) (F1-macro) | | | | End-to-End (%) (F1-macro) | | | |
|---|---|---|---|---|---|---|---|---|---|---|---|---|
| | MGB-Med | Re-CASI | MIV-Med | Avg | MGB-Med | Re-CASI | MIV-Med | Avg | MGB-Med | Re-CASI | MIV-Med | Avg |
| GPT-4o | **88.2 ± 0.3** | **96.1 ± 0.3** | 90.0 ± 0.2 | **91.4** | **89.1 ± 0.4** | **86.6 ± 0.6** | 68.4 ± 0.5 | 81.4 | **77.5 ± 1.1** | **80.4 ± 1.3** | 64.1 ± 0.2 | **74.0** |
| Llama-3.1-8B-Instruct | 47.3 ± 0.3 | 94.5 ± 0.2 | 89.2 ± 0.1 | 77.0 | 59.2 ± 0.3 | 68.4 ± 0.7 | 66.0 ± 1.1 | 64.5 | 40.6 ± 0.4 | 58.2 ± 0.4 | 61.9 ± 2.8 | 53.6 |
| Llama-3.1-70B-Instruct | 77.3 ± 0.8 | 95.7 ± 0.3 | **90.1 ± 0.1** | 87.7 | 88.4 ± 0.4 | 85.4 ± 1.0 | 73.7 ± 1.7 | 82.5 | 67.9 ± 1.1 | 79.4 ± 0.3 | **66.8 ± 0.5** | 71.4 |

| Model | Medication Extraction (%) (F1-micro) | | | | Medication Status Classification (%) (F1-macro) | | | | End-to-End (%) (F1-macro) | | | |
|---|---|---|---|---|---|---|---|---|---|---|---|---|
| | MGB-Med | Re-CASI | MIV-Med | Avg | MGB-Med | Re-CASI | MIV-Med | Avg | MGB-Med | Re-CASI | MIV-Med | Avg |
| Meditron-7b | 39.0 ± 0.4 | 84.5 ± 0.4 | 76.2 ± 0.5 | 66.6 | 48.7 ± 1.1 | 49.6 ± 0.8 | 49.3 ± 0.2 | 49.2 | 22.2 ± 1.1 | 46.9 ± 0.3 | 43.2 ± 0.3 | 37.4 |
| Meditron-70b | 41.5 ± 0.2 | 93.0 ± 0.3 | 86.4 ± 0.2 | 73.6 | 69.6 ± 1.0 | 60.1 ± 1.0 | 60.6 ± 1.9 | 63.4 | 35.8 ± 0.4 | 57.4 ± 0.4 | 55.0 ± 1.5 | 49.4 |
| MeLLaMA-13B-chat | 42.8 ± 0.7 | 91.1 ± 0.3 | 82.7 ± 0.3 | 72.2 | 64.7 ± 0.5 | 58.9 ± 0.9 | 58.3 ± 0.3 | 60.6 | 30.8 ± 0.6 | 55.9 ± 0.3 | 48.8 ± 0.7 | 45.2 |
| MeLLaMA-70B-chat | 58.6 ± 0.4 | 93.4 ± 0.2 | 87.3 ± 0.4 | 79.8 | 88.2 ± 0.6 | 67.7 ± 0.9 | 61.6 ± 1.4 | 72.5 | 52.1 ± 0.7 | 60.7 ± 1.8 | 54.0 ± 0.7 | 55.6 |
| Mistral-7B-Instruct-v0.3 | 37.0 ± 0.1 | 86.8 ± 0.5 | 85.3 ± 0.2 | 69.7 | 69.0 ± 0.8 | 70.7 ± 0.4 | 69.7 ± 0.9 | 69.8 | 31.8 ± 0.4 | 59.3 ± 0.5 | 52.5 ± 0.6 | 47.9 |
| Mistral-Nemo-Instruct-2407 | 46.6 ± 0.3 | 93.3 ± 0.2 | 87.8 ± 0.2 | 75.9 | 79.2 ± 0.6 | 78.7 ± 0.8 | 75.8 ± 0.5 | 77.9 | 41.1 ± 0.3 | 67.4 ± 0.4 | 62.5 ± 0.3 | 57.0 |
| Qwen2.5-14B-Instruct | 71.0 ± 0.9 | 94.6 ± 0.1 | 88.9 ± 0.2 | 84.8 | 85.0 ± 0.4 | 78.9 ± 0.8 | 68.3 ± 1.9 | 77.4 | 61.2 ± 0.6 | 71.8 ± 0.3 | 58.7 ± 0.5 | 63.9 |
| Qwen2.5-32B-Instruct | 88.5 ± 0.5 | 95.0 ± 0.1 | 88.5 ± 0.1 | 90.7 | 90.7 ± 0.0 | 83.3 ± 0.2 | 68.9 ± 0.4 | 81.0 | 78.1 ± 0.7 | 75.2 ± 0.3 | 65.1 ± 0.6 | 72.8 |
| Qwen2.5-72B-Instruct | 70.2 ± 0.6 | 95.3 ± 0.1 | 87.9 ± 0.0 | 84.5 | 87.8 ± 0.1 | 84.7 ± 0.1 | **75.9 ± 0.7** | **82.8** | 61.1 ± 0.6 | 77.8 ± 0.1 | 66.4 ± 0.3 | 68.4 |

**Table 4**. Model Performance under CoT Prompting

| Model | Medication Extraction (%) (F1-micro) | | | | Medication Status Classification (%) (F1-macro) | | | | End-to-End (%) (F1-macro) | | | |
|---|---|---|---|---|---|---|---|---|---|---|---|---|
| | MGB-Med | Re-CASI | MIV-Med | Avg | MGB-Med | Re-CASI | MIV-Med | Avg | MGB-Med | Re-CASI | MIV-Med | Avg |
| GPT-4o | **96.8 ± 0.6** | 93.8 ± 0.3 | **89.2 ± 0.1** | **93.3** | 85.4 ± 0.8 | **85.0 ± 1.1** | 65.4 ± 0.6 | **78.6** | **83.1 ± 1.0** | 75.7 ± 1.0 | **60.8 ± 0.6** | **73.2** |
| Llama-3.1-8B-Instruct | 29.5 ± 0.7 | 89.0 ± 0.4 | 82.9 ± 0.8 | 67.1 | 51.5 ± 1.0 | 63.1 ± 0.7 | 57.6 ± 1.3 | 57.4 | 19.9 ± 0.6 | 58.1 ± 0.8 | 50.7 ± 0.4 | 42.9 |
| Llama-3.1-70B-Instruct | 76.2 ± 1.6 | 94.0 ± 0.3 | 86.7 ± 0.4 | 85.6 | 84.2 ± 1.7 | 79.0 ± 1.4 | 63.9 ± 0.9 | 75.7 | 66.1 ± 1.8 | **77.6 ± 2.4** | 57.9 ± 0.5 | 67.2 |
| Meditron-7b | 21.8 ± 0.1 | 45.4 ± 2.6 | 39.6 ± 1.1 | 35.6 | 0.0 ± 0.0 | 4.2 ± 0.3 | 5.6 ± 0.5 | 3.3 | 12.4 ± 0.4 | 20.8 ± 1.7 | 20.7 ± 0.8 | 18.0 |
| Meditron-70b | 33.7 ± 0.4 | 91.4 ± 2.3 | 74.9 ± 1.8 | 66.7 | 60.2 ± 2.3 | 68.3 ± 0.8 | 57.0 ± 0.2 | 61.8 | 19.9 ± 0.5 | 57.3 ± 1.9 | 44. 0 ± 2.0 | 40.4 |
| MeLLaMA-13B-chat | 19.9 ± 0.1 | 87.1 ± 0.6 | 74.6 ± 0.9 | 60.5 | 22.1 ± 0.6 | 62.1 ± 0.5 | 53.4 ± 0.7 | 45.9 | 17.5 ± 0.5 | 55.4 ± 0.9 | 40.7 ± 0.6 | 37.9 |
| MeLLaMA-70B-chat | 52.9 ± 2.1 | 90.7 ± 0.9 | 77.9 ± 1.3 | 73.8 | **87.1 ± 0.7** | 78.2 ± 1.9 | 59.7 ± 0.7 | 75.0 | 41.9 ± 1.6 | 62.7 ± 0.7 | 53.3 ± 2.7 | 52.6 |
| Mistral-7B-Instruct-v0.3 | 53.2 ± 0.5 | 82.2 ± 0.7 | 74.0 ± 0.6 | 69.8 | 75.8 ± 0.5 | 59.8 ± 1.0 | 51.0 ± 0.6 | 62.2 | 43.7 ± 0.7 | 58.2 ± 0.7 | 40.3 ± 0.3 | 47.4 |
| Mistral-Nemo-Instruct-2407 | 52.5 ± 0.6 | 89.1 ± 0.8 | 83.2 ± 0.6 | 74.9 | 72.7 ± 0.7 | 79.6 ± 2.0 | 59.6 ± 0.7 | 70.6 | 43.6 ± 0.8 | 68.5 ± 1.7 | 52.7 ± 1.3 | 54.9 |

| Model | Medication Extraction (%) (F1-micro) | | | | Medication Status Classification (%) (F1-macro) | | | | End-to-End (%) (F1-macro) | | | |
|---|---|---|---|---|---|---|---|---|---|---|---|---|
| | MGB-Med | Re-CASI | MIV-Med | Avg | MGB-Med | Re-CASI | MIV-Med | Avg | MGB-Med | Re-CASI | MIV-Med | Avg |
| Qwen2.5-14B-Instruct | 64.5 ± 1.8 | 93.5 ± 0.8 | 87.7 ± 0.3 | 81.9 | 87.7 ± 0.2 | 83.6 ± 1.2 | **66.8 ± 1.3** | 79.4 | 52.5 ± 1.2 | 69.7 ± 1.6 | 58.3 ± 0.6 | 60.2 |
| Qwen2.5-32B-Instruct | 81.9 ± 1.8 | **94.5 ± 0.9** | 84.7 ± 0.6 | 87.0 | 85.2 ± 1.5 | 83.6 ± 1.0 | 65.0 ± 0.4 | 77.9 | 72.2 ± 2.3 | 76.9 ± 0.9 | 54.0 ± 1.5 | 67.7 |
| Qwen2.5-72B-Instruct | 93.1 ± 0.6 | 91.6 ± 0.9 | 84.4 ± 0.3 | 89.7 | 83.2 ± 0.6 | 80.7 ± 0.9 | 66.8 ± 1.5 | 76.9 | 74.0 ± 0.6 | 73.5 ± 1.0 | 56.1 ± 1.2 | 67.9 |

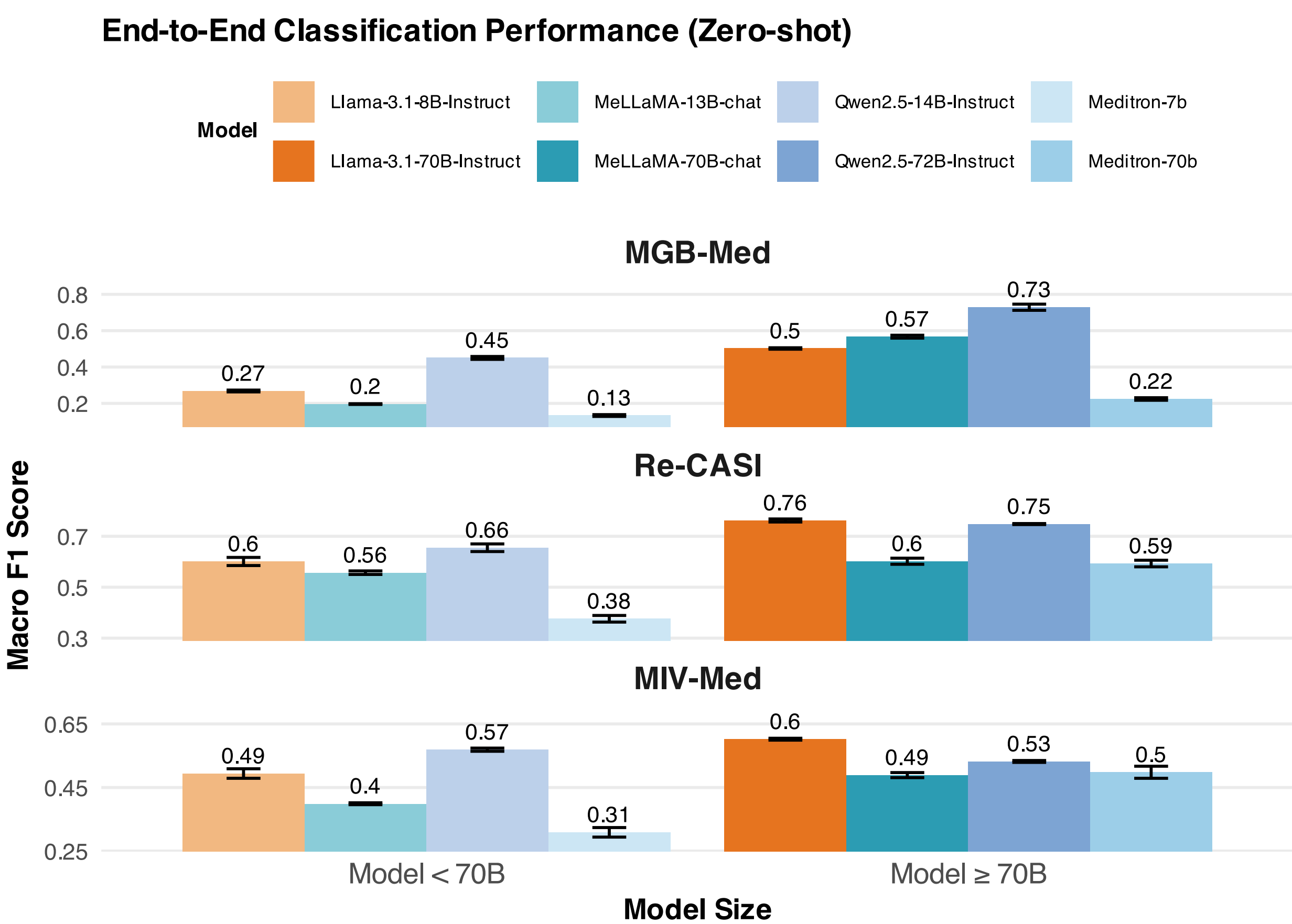

**Figure 3:** Zero-shot Performance Comparison of Small (< 70B) and Large (≥ 70B) Open-sourced Models for the End-to-End Classification Task. Error bars represent the standard deviation.

### 4.2 Small LLM (< 70B) and large LLM (≥ 70B) comparison

To assess performance across LLM sizes, we compared four LLM suites (Llama 3.1, Me-LLaMA, Qwen 2.5, Meditron), each consisting of a smaller model (7–13B) and a larger model (~70B) on the end-to-end task in Figure 3. Across all datasets, large models (≥ 70B parameters) generally outperformed small models (< 70B parameters). This finding underscores that scaling improves LLMs' performance in clinical text processing[38].

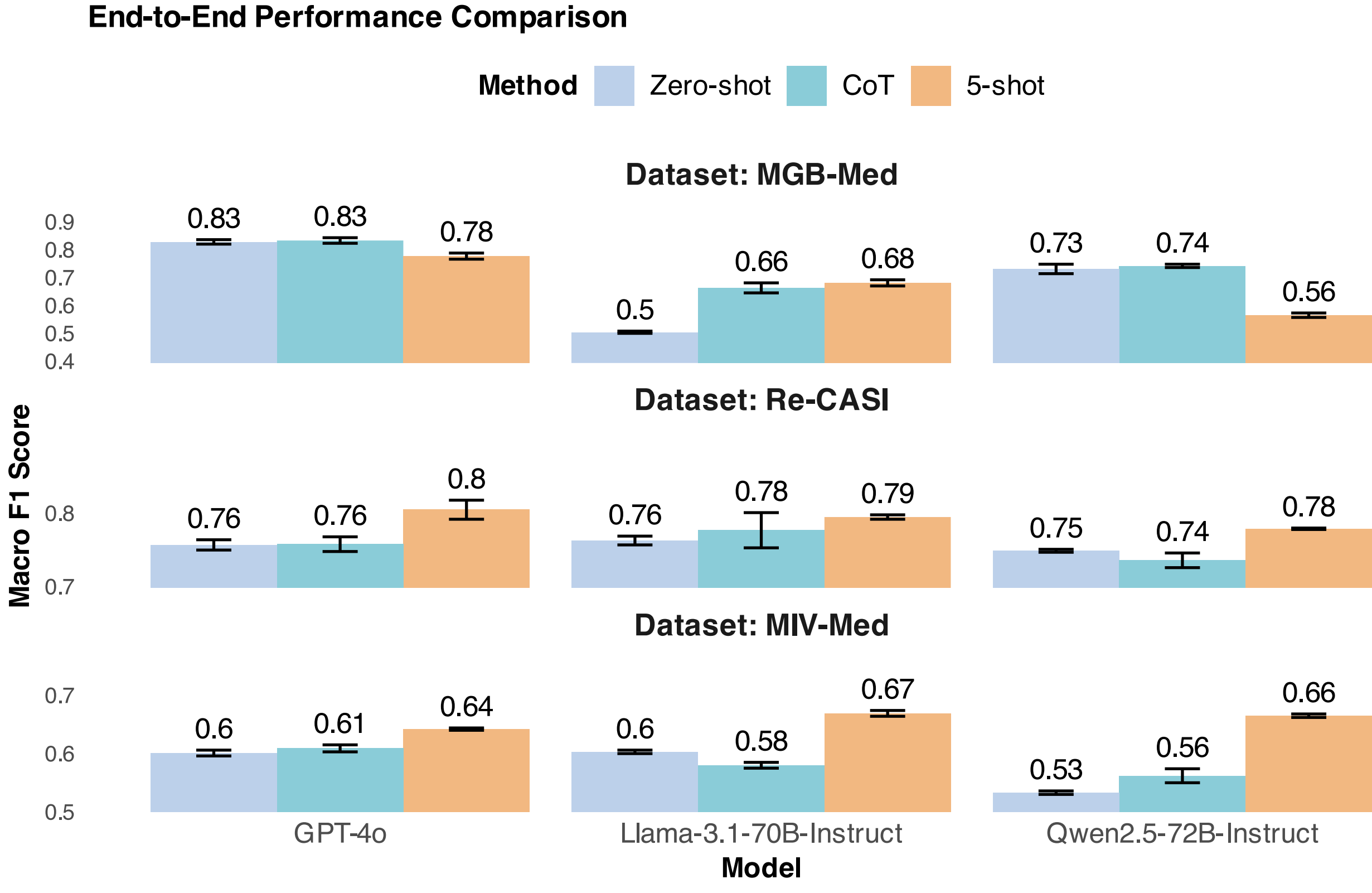

**Figure 4:** Performance Comparison of GPT-4o, Llama-3.1-70B-Instruct, and Qwen2.5-72B-Instruct on the End-to-End Classification Task. Error bars represent the standard deviation.

### 4.3 Inference strategies comparison

We compared inference strategies using the top three models (GPT-4o, Llama-3.1-70B-Instruct, and Qwen2.5-72B-Instruct) on the end-to-end task across three datasets (Figure 4). Compared with zero-shot, 5-shot prompting often improved performance, indicating that examples help the models capture task-specific nuances in clinical text. However, in MGB-Med, 5-shot prompting underperformed. Because MGB-Med omits the "Neither" label, borderline constructions (e.g., "hold," "resume," taper/PRN directives, or allergy/"avoid" notes) were forced into "active" or "discontinued" in the few-shot examples, which likely introduced inconsistency and hindered generalization. In contrast, the other two datasets, which used clearer and more consistent prompts, demonstrated gains with 5-shot prompting. CoT effects were inconsistent: Llama-3.1-70B-Instruct improved on MGB-Med; GPT-4o showed negligible differences between zero-shot and CoT; and CoT slightly reduced performance in some cases (Llama-3.1-70B-Instruct on MIV-Med; Qwen2.5-72B-Instruct on Re-CASI).

### 4.4 General and medical domain LLMs comparison

Medical-specific LLMs such as Me-LLaMA and Meditron models, which were trained on medical-related data, were expected to have an advantage, but did not surpass top-tier general-purpose models like Llama3.1-70B-Instruct or Qwen2.5-72B-Instruct (Table 2-4). Both Me-LLaMA and Meditron model suites were fine-tuned on Llama 2 models[13,14], which are relatively outdated and performed lower than the newer, general-domain Llama 3.1[10]. Me-LLaMA models consistently outperformed Meditron models across all tasks and datasets, likely reflecting training differences. Meditron models were trained primarily on clinical guidelines, medical literature, and general-domain texts[14]—not on EHRs. In contrast, Me-LLaMA models were continually pre-trained on de-identified EHR notes (MIMIC-III and MIMIC-IV), biomedical literature, and general-domain data[13], making them better suited for tasks on EHR notes. These results highlight the value of EHR-specific pretraining for real-world clinical text understanding.

### 4.5 Error Analysis

We selected GPT-4o and Llama-3.1-70B-Instruct, the top-performing models from the proprietary and open-source categories, respectively, for error analysis. We visualized confusion matrices for medication status classification in Figure 5. The row totals (true labels) are identical across models and match the ground truth distributions reported in Table 1. Because LLMs might generate results that fall outside predefined medication status categories, fail to comply with the response format, or omit information recorded in the EHR, we categorize such outputs as "invalid" in Figure 5. Importantly, "Invalid" is not a ground truth label and reflects model-specific hallucinations or schema violations; thus it legitimately differs across models. The true label distributions remain identical across models, as confirmed by the row totals in Figure 5, which match the counts reported in Table 1 (e.g., MGB-Med: 327 Active, 116 Discontinued). For example, the "Invalid" category in the true label axis refers to medications that were not present in the EHR snippets but were nonetheless assigned discontinuation status labels by LLMs. Conversely, the "Invalid" category in the prediction label axis refers to cases where the LLM output does not identify the discontinuation status for a medication that was mentioned in the EHR. Both models correctly identified most "Active" and "Discontinued" cases in Re-CASI and MIV-Med datasets. GPT-4o produced fewer invalid predictions than Llama-3.1-70B-Instruct, especially in the MGB-Med and MIV-Med datasets. In MGB-Med, Llama-3.1-70B-Instruct generated 622 spurious "Active" classifications and 135 "Discontinued" classifications (170.9% relative to dataset size). In contrast, GPT-4o only had 1 spurious "Active" and 1 "Discontinued" classification. This indicates a tendency for Llama-3.1-70B-Instruct to hallucinate medications where none exist according to the ground truth. Across all datasets, both models struggled with the "Neither" category, where notes reference medications without an explicit action. As illustrated in Appendix A.1, "Neither" commonly reflects avoidance recommendations (Ex. 3: ibuprofen, Naprosyn), hypothetical/contingent plans (Ex. 4: Duragesic, MS Contin), and historical or declined use (Ex. 1: Celexa). This underscores the challenges LLMs face when interpreting contextual ambiguity and implicit references in medication instructions.

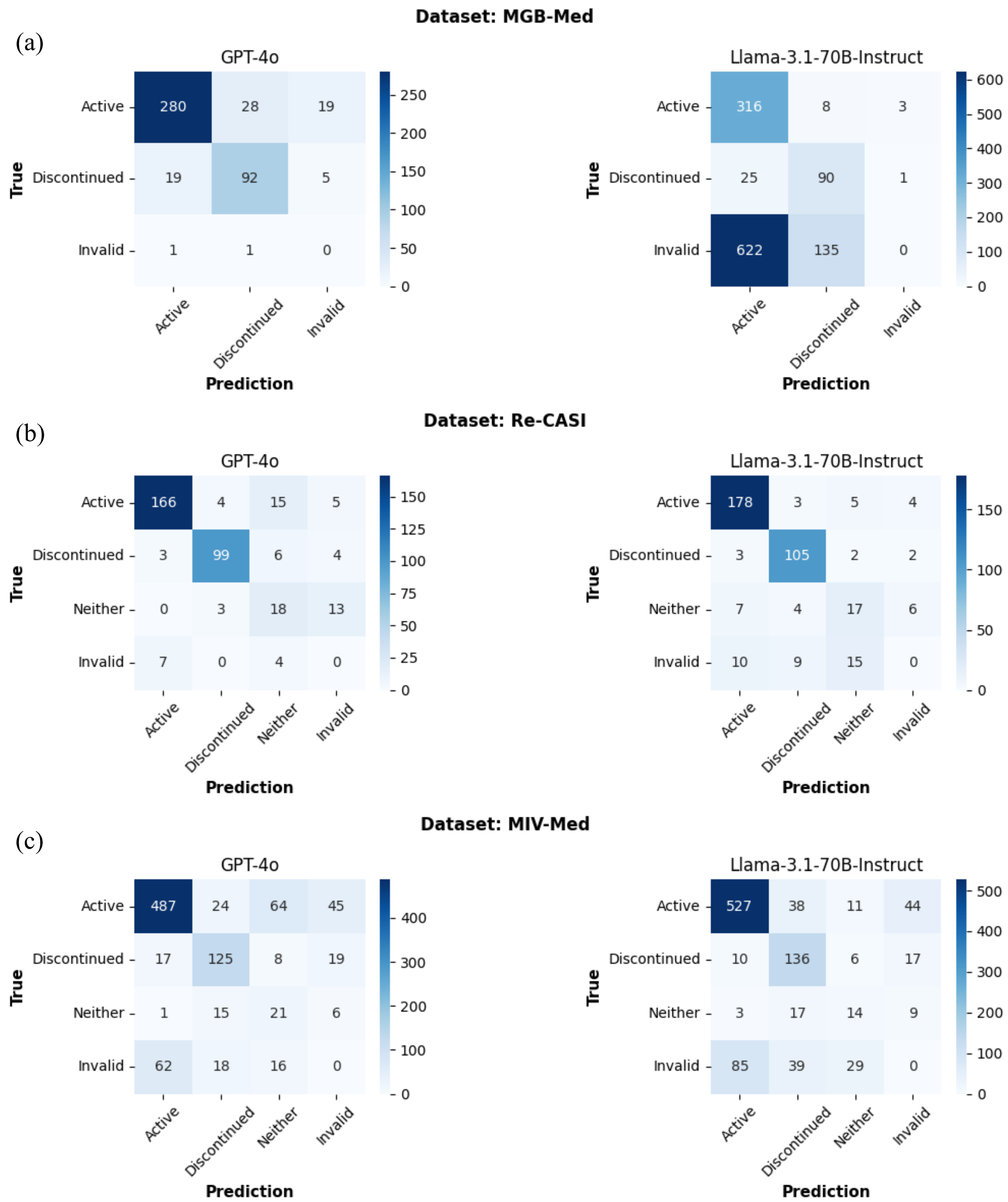

**Figure 5:** Confusion matrix of GPT-4o and Llama-3.1-70B-Instructon on medication status classification using Zero-shot prompting.

Table 5: Invalid predictions (hallucination rate) by dataset and model using Zero-shot prompting.

| Dataset | Total medications | GPT-4o invalid count (%) | Llama-3.1-70B Invalid count (%) |
|---|---|---|---|
| MGB-Med | 443 (327 Active, 116 Discontinued) | 2 (0.45%) | 757 (170.88%) |
| Re-CASI | 336 (190 Active, 112 Discontinued, 34 Neither) | 11(3.27%) | 34(10.12%) |
| MIV-Med | 832 (620 Active, 169 Discontinued, 43 Neither) | 96(11.54%) | 153(18.39%) |

As shown in Table 5, GPT-4o consistently produced low invalid rates across all datasets (0.45% in MGB-Med, 3.27% in Re-CASI, and 11.54% in MIV-Med). By contrast, Llama-3.1-70B-Instruct generated substantially more invalid predictions, with 10.12% in Re-CASI, 18.39% in MIV-Med, and an inflated count in MGB-Med (757 invalid predictions, equivalent to 170.88% relative to dataset size). Across datasets, spurious medication mentions often arise when models process avoidance instructions (e.g., “avoid ibuprofen”), hypothetical treatment plans, or historical references. These contexts were frequently misclassified as “active” or “discontinued” rather than “neither.” Another contributing factor is dataset structure: in the MGB-Med dataset, snippets are considerably longer than in the Re-CASI or MIV-Med datasets (91 vs. 63–68 words on average), and individual medications are sometimes mentioned multiple times with changing statuses (e.g., from active to discontinued or resumed later). This increases the likelihood of inconsistent outputs and hallucinated labels. To mitigate such errors, future research should investigate instruction tuning specifically adapted for medication extraction and status classification. Recent large-scale benchmarking shows task-specific tuning consistently improves robustness and reduce errors, including hallucinations, across a range of biomedical NLP tasks[39]. This suggests a principled strategy for enhancing model reliability in medication status classification.

## 5.0 Discussion

This study highlights the growing potential of LLMs in clinical informatics, specifically in automating medication extraction and status classification from diverse clinical texts. State-of-the-art LLMs, including GPT-4o and open-source models like Qwen2.5-72B-Instruct and Llama-3.1-70B-Instruct, can achieve impressive performance on these tasks.

Our experiments show that few-shot prompting more consistently improved performance than CoT prompting. First, few-shot examples explicitly demonstrate what to extract and how to classify statuses. Second, they promote schema adherence by guiding outputs to the required format. In contrast, CoT prompting encourages step-by-step reasoning, which, without comprehensive retained medical knowledge, may increase the risk of LLM hallucination and incorrect inferences[40]. Previous studies also observed worse CoT predictions compared with zero-shot on clinical text understanding[37,41–43]. A detailed exploration of CoT limitations is deferred to future work.

Beyond model performance, automating large-scale medication extraction and status classification has broad implications for clinical research and healthcare delivery. Pharmacoepidemiology can leverage LLM-extracted features to enhance drug utilization monitoring, adherence tracking, and adverse effect detection through large-scale data analyses[44]. When linked with patient outcomes and disease history, such structured information could support longitudinal studies on treatment effectiveness and safety. Additionally, integrating extracted data into disease registries and clinical notes could help build medication knowledge graphs, enabling better clinical decision support and predictive analytics for precision medicine[44].

Interestingly, medical-specific models showed no clear advantage over general-purpose models, as recent general-purpose open-source models (e.g., Llama-3.1-70B-Instruct and Qwen2.5-72B-Instruct) consistently outperformed Me-LLaMA and Meditron models. This challenges the assumption that domain-specific pretraining necessarily yields superior performance in clinical NLP tasks. One possible reason is the limitation of the relatively outdated base models, as Me-LLaMA models were trained on Llama 2[13], behind Llama 3.1[10]. Another contributing factor is the limited size and scope of the training data: Me-LLaMA was trained on the limited MIMIC EHR corpus, while Meditron was pre-trained on PubMed but not real-world EHR notes and lacked specific instruction-tuning. Access to large-scale EHR notes remains restricted[45], which limits the development and training of EHR-specific LLMs. One future direction is to build on advanced LLMs and introduce additional medical knowledge, either through fine-tuning or retrieval-augmented generation[41,46] on large-scale real-world EHR notes, to further enhance the capabilities of existing high-performance general-domain LLMs in understanding EHRs.

This study has several limitations. First, datasets were curated to include only snippets that contain medication information rather than entire EHR notes, which may oversimplify real-world complexities and limit generalizability. Although snippets were selected at the complete-sentence level to preserve intra-sentence temporal cues, this may omit the cross-sentence/section/encounter context required for temporal reasoning. Second, this study primarily focuses on evaluating LLM performance for medication-related tasks under common inference settings. Advanced strategies such as instruction tuning, improved few-shot example selection, and retrieval-augmented generation (RAG) were not included and will be explored in future work. Third, due to computational constraints and institutional access limitations, recently developed reasoning LLMs (e.g., DeepSeek-R1) were not incorporated. These models, which generate explicit reasoning traces, may offer valuable insights for understanding model behavior and facilitating more detailed error analysis. Finally, for privacy and data security reasons, the MGB and MIMIC-IV datasets are not publicly available. Therefore, a fixed set of few-shot examples from the publicly accessible Re-CASI dataset was used to support demonstration, analysis, and reproducibility. While this approach may introduce dataset bias, the consistent performance improvements observed across tasks support the robustness of our findings.

## 6.0 Conclusions

This study demonstrates the potential of LLMs for extracting medication mentions and classifying status from EHR notes. In some settings, open-source models, such as Llama-3.1-70B-Instruct and Qwen2.5-72B-Instruct, performed comparably with proprietary systems like GPT-4o, offering avenues for privacy-compliant, scalable research tools. While results varied across datasets and prompting strategies, our findings indicate that LLMs can extract and classify medications from unstructured clinical text, offering scalable and privacy-compliant alternatives to enhance pharmacovigilance and clinical decision support in real-world healthcare settings.

## 7.0 Acknowledgments

This work was funded by PCORI ME-2022C1-25646, Goldberg Scholarship, Brigham Research Institute, National Institute on Aging RF1AG090405, and National Library of Medicine R01LM014667.

## 8.0 Data Availability

The Re-CASI dataset used in this study is publicly available and can be accessed at this Hugging Face repository (https://huggingface.co/datasets/mitclinicalml/clinical-ie/tree/main).

The MIV-Med dataset is derived from the MIMIC IV and annotated by our team. MIMIC IV is publicly accessible through PhysioNet (https://physionet.org) and the annotated MIV-Med dataset will be available in PhysioNet once this paper is published.

The MGB-Med dataset was annotated using internal clinical notes from Mass General Brigham and contains protected health information. To protect patient privacy, this dataset is not publicly available.

## 9.0 Declaration of generative AI and AI-assisted technologies in the writing process

During the preparation of this work, the authors used GPT-4o for grammar checking. After using this tool, the authors reviewed and edited the content as needed and take full responsibility for the final content of the publication. No AI tools were used for study design, data collection, or data analysis.

# 11.0 Appendix

## A.1 Annotation guidelines for MIMIC-IV discharge summary snippets

*1. Introduction*

This document outlines the guidelines for annotating medication mentions in discharge summaries from the MIMIC-IV dataset. The goal of the annotation task is to create snippets from MIMIC-IV discharge summaries, then extract medications into three categories:

- Active (Current) Medications: Medications the patient is currently taking.
- Discontinued Medications: Medications the patient previously took but has stopped.
- Other Mentioned Medications: Medications mentioned that are neither currently taken nor discontinued (e.g., historical or comparative mentions).

The correct identification of medications is critical for research and analysis of patient treatment patterns.

*2. Annotation Tool and Process*

*2.1 Snippet creation process*

The snippets for annotation were derived from the MIMIC-IV discharge summaries dataset. The process for creating the snippets was as follows:

1. Selection of Discharge Summaries: A random set of discharge summaries was selected from the MIMIC-IV Notes dataset.
2. Sentence Splitting: Each selected discharge summary was split into individual sentences using the PyRush sentence segmentation tool[47].

3. Initial Drug Mention Identification: Sentences containing potential drug mentions were identified using keyword matching from RxNorm, a standardized nomenclature for clinical drugs[35].
4. Snippet Formation: For each sentence containing a drug mention, two preceding and two following sentences were added to provide context. These sets of contiguous sentences formed a snippet of text.
5. Snippet Review: Snippets were further reviewed manually using a spreadsheet application. During this step, the following criteria were applied:
   - Exclusion of irrelevant snippets: Snippets that did not contain an actual mention of a drug (e.g., false positives from keyword matching) were excluded.
   - Exclusion of purely medication lists: Snippets that were only lists of medications, without any accompanying narrative text, were also excluded. Snippets had to contain at least some narrative to be considered for annotation.

*2.2 Annotation process*

Once the snippets were finalized, the annotation process was conducted using a simple spreadsheet application (e.g., LibreOffice or Excel) without the use of specialized annotation tools. The annotation process involved the following steps:

1. Reading the Snippet: Annotators carefully read each snippet to identify medications mentioned in the text.
2. Categorization of Medications: Annotators categorized each identified medication into one of three categories:
   - Active Medications: Medications that the patient is currently taking at the time of discharge.

- Discontinued Medications: Medications that the patient was previously prescribed but has since stopped taking.
   - Neither Medications: Medications mentioned in the snippet that are neither active nor discontinued (e.g., historical mentions, medications discussed but not prescribed).
3. Recording Annotations: The categorized medications were recorded in a comma-separated list in the appropriate columns of the spreadsheet, corresponding to "Active Medications," "Discontinued Medications," and "Neither Medications."
4. Output Format: The final annotations adhered to a specific output format, ensuring consistency across all annotated snippets. For each snippet, medications were listed under the appropriate category in a comma-separated format.

*3. Annotation Specification*

In this task, annotators will categorize each medication mentioned in the snippets into one of three categories: Active Medications (Current), Discontinued Medications, or Neither Medications.

*3.1. Active medications (Current)*

- Definition: These are medications that the patient is currently taking or has been newly prescribed during the course of treatment mentioned in the snippet. The use of the medication is ongoing at the time of the snippet.
- Identification: Look for indications that the patient is continuing or starting the medication. Phrases like "start," "continue," or "currently taking" are strong indicators.

*3.2. Discontinued medications*

- Definition: Medications that the patient was previously taking but has since stopped. This includes medications that have been explicitly discontinued or placed on hold by the healthcare provider.
- Identification: Look for language that suggests a medication is no longer being used, such as "discontinued," "stopped," or "held."

*3.3. Neither medications*

- Definition: Medications mentioned that do not fall into the categories of "Active" or "Discontinued." These may include:
  - Historical medications that the patient has taken in the past but is not currently on.
  - Medications listed as allergies or sensitivities.
  - Medications discussed or considered but not prescribed.
  - Antibiotics or drugs where resistance or sensitivity is noted.

## *4. Example Annotations*

Due to restrictions on publishing text from the MIMIC-IV dataset, the following examples are taken from the publicly available CASI dataset, which served as a template for creating the current dataset. These examples illustrate how to annotate snippets based on the active, discontinued, and neither medication categories, as outlined in the Annotation Specification.

### Example 1

Snippet: "I will recommend discontinuing the alcohol withdrawal protocol and start her on Ativan 1 mg p.o. q. 8 hours and use Ativan 1 mg IV q. 4 hours p.r.n. for agitation. I will also start her on Inderal LA 60 mg p.o. q.d. for essential tremors. She does not want to take Celexa, and I will put her back on Lexapro 2 mg p.o. q.d."

- Active Medications: Lexapro, Ativan, Inderal LA
- Discontinued Medications: [None]
- Neither Medications: Celexa

### Example 2

Snippet: "The patient remained afebrile for the last 72 hours. Plan was made to complete the course of linezolid and Levaquin. Considering his immunosuppressed status, the plan is to initially complete the therapeutic course of Bactrim for PCP and then will continue patient on Bactrim prophylaxis dose. Will also continue patient on prophylactic antifungal regime. Given that the patient has MRSA pneumonia we thought that the patient might also have MRSA sinusitis, and we switched the patient from vancomycin to linezolid since linezolid has better soft tissue penetration."

- Active Medications: linezolid, Levaquin, Bactrim
- Discontinued Medications: vancomycin
- Neither Medications: [None]

### Example 3

Snippet: "Her aspirin (81 mg q.d.) is discontinued, and the patient is advised that she needs to avoid ibuprofen, Naprosyn, alcohol, caffeine, and chocolate. She is advised that Tylenol 325 mg or Tylenol ES (500 mg) is safe to take at 1 or 2 q.4-6h. p.r.n. for pain or fever. DISCHARGE MEDICATIONS: Plavix 75 mg p.o. q.d."

- Active Medications: Plavix, Tylenol, Tylenol ES
- Discontinued Medications: aspirin
- Neither Medications: ibuprofen, Naprosyn

### Example 4

Snippet: "She will be started on Tequin to treat her urinary tract infection and may consider switching IV Morphine again to MS Contin, although, she may be having nausea secondary to the pain or the MS Contin. If she does not tolerate oral Morphine, Duragesic patch will be reasonable based on her multiple allergies to other narcotic analgesics."

- Active Medications: Morphine, Tequin
- Discontinued Medications: [None]
- Neither Medications: Duragesic, MS Contin

## A.2 Annotator agreement analysis

*Computing F1 Score for the Extraction Task*

The inter-annotator agreement for the extraction task was evaluated using the F1 score, a widely adopted metric for binary classification tasks. The process involved first identifying all medications mentioned by each annotator for every snippet in the dataset. For Annotator 1, medications labeled as "active," "discontinued," or "neither" were consolidated into a single set. The same procedure was applied to Annotator 2, ensuring a comprehensive representation of their respective extractions.

Subsequently, a unified set of all medications mentioned by either annotator was created for each snippet. For every medication in this unified set, binary encodings were generated to denote whether the medication was extracted by a given annotator. Specifically, a value of 1 was assigned if the annotator extracted the medication, and 0 otherwise. This process yielded two binary arrays, corresponding to the medications extracted by Annotator 1 and Annotator 2, respectively.

Using these binary encodings, precision, recall, and the F1 score were calculated. Precision was defined as the ratio of true positives (medications correctly identified by both annotators) to the total number of medications extracted by Annotator 1. Recall was computed as the ratio of true positives to the total number of medications identified by Annotator 2. Finally, the F1 score, representing the harmonic mean of precision and recall, was calculated to provide a balanced metric for the annotators' agreement in the extraction task.

*Computing Cohen's Kappa score for the classification task*

Cohen's Kappa score was measured to assess the inter-annotator agreement for the classification tasks. This metric quantifies agreement between annotators while accounting for agreement expected by chance. The process begins by considering each snippet and identifying the union of all medications extracted by both annotators. This ensures that every medication mentioned by either annotator is included in the analysis.

For every medication in this union, its classification status was determined based on its presence in one of three predefined categories: "active", "discontinued", or "neither". The classification assigned by Annotator 1 was derived from their respective categorizations, and the same procedure was followed for Annotator 2. The resulting classifications for each medication were appended to separate lists corresponding to the two annotators.

To facilitate the computation of Cohen's Kappa, the categorical classifications ("active," "discontinued," and "neither") were mapped to numerical values: 1 for "active," 2 for "discontinued," and 0 for "neither." The Kappa score was subsequently calculated using these numerical representations, providing a robust measure of agreement between the annotators. A high Kappa score reflects strong concordance, emphasizing the consistency of annotator classifications in the task.

### A.3 Example prompt

For MGB-Med, the prompt template was identical but the "Neither" option was omitted to match the dataset; example text from MGB-Med cannot be shown due to PHI, so CASI examples are displayed.

Supplementary Table 1: Medication Extraction and End-to-end Task Prompts.

| Zero-shot Prompt | CoT Prompt | 5-shot Prompt |
| --- | --- | --- |
| Input Medical Note:{}<br>Create a bulleted list of mentioned medications and classify whether they are in the status of active, discontinued, or neither.<br>Expected output format:<br>- medication_name (status)<br>- ...<br>END<br><br>Output: | Input Medical Note: {}<br>Create a bulleted list of mentioned medications and classify whether they are in the status of active, discontinued, or neither.<br>Let's think step by step and then give the answer in the expected format at the end.<br>Expected output format:<br>Step-by-Step Thoughts:<br>1. [thought]<br>2. ...<br><br>Answer:<br>- medication_name (status)<br>- ...<br>END<br>Output: | Input Medical Note: _%#NAME#%_ tolerated his chemotherapy well with no significant side effects. He was given 6 MP orally daily and Bactrim for prophylaxis. He was on dapsone in the past.<br>Create a bulleted list of mentioned medications and classify whether they are in the status of active, discontinued, or neither.<br>Expected output format:<br>- medication_name (status)<br>- ...<br>END<br>Output:<br>- 6 MP (active)<br>- Bactrim (active)<br>- dapsone (discontinued)<br>END<br>Input Medical Note: {}<br>Create a bulleted list of mentioned medications and classify whether they are in the status of active, discontinued, or neither.<br>Expected output format:<br>- medication_name (status)<br>- ...<br>END<br><br>Output: |

Supplementary Table 2: Medication Status Classification Task Prompts.

| Zero-shot Prompt | CoT Prompt | 5-shot Prompt |
|---|---|---|
| Input Medical Note:{}<br>Create a bulleted list of mentioned medications and classify whether they are in the status of active, discontinued, or neither.<br>Expected output format:<br>- medication_name (status)<br>- ...<br>END<br><br>Hint: Here is a complete list of medications included in this note: {}. Assign a status for each of them.<br>Output: | Input Medical Note: {}<br>Create a bulleted list of mentioned medications and classify whether they are in the status of active, discontinued, or neither.<br>Let's think step by step and then give the answer in the expected format at the end.<br>Expected output format:<br>Step-by-Step Thoughts:<br>1. [thought]<br>2. ...<br><br>Answer:<br>- medication_name (status)<br>- ...<br>END<br><br>Hint: Here is a complete list of medications included in this note: {}. Assign a status for each of them.<br>Output: | Input Medical Note: _%#NAME#%_ tolerated his chemotherapy well with no significant side effects. He was given 6 MP orally daily and Bactrim for prophylaxis. He was on dapsone in the past.<br>Create a bulleted list of mentioned medications and classify whether they are in the status of active, discontinued, or neither.<br>Expected output format:<br>- medication_name (status)<br>- ...<br>END<br><br>Hint: Here is a complete list of medications included in this note: 6 MP, Bactrim, dapsone. Assign a status for each of them.<br>Output:<br>- 6 MP (active)<br>- Bactrim (active)<br>- dapsone (discontinued)<br>END<br><br>Input Medical Note: {}<br>Create a bulleted list of mentioned medications and classify whether they are in the status of active, discontinued, or neither.<br>Expected output format:<br>- medication_name (status)<br>- ...<br>END<br><br>Hint: Here is a complete list of medications included in this note: {}. Assign a status for each of them.<br>Output: |